\documentclass[runningheads]{llncs}
\usepackage[T1]{fontenc}
\usepackage{graphicx}
\usepackage{adjustbox}
\begin{document}
\title{Metric Learning and Adaptive Boundary for Out-of-Domain Detection}
\titlerunning{Metric Learning and Adaptive Boundary for Out-of-Domain Detection}
%
\author{Petr Lorenc\inst{1} \and
Tommaso Gargiani\inst{1} \and
Jan Pichl\inst{1} \and
Jakub Konrád\inst{1} \and
Petr Marek\inst{1} \and
Ondřej Kobza\inst{1} \and
Jan Šedivý\inst{2}
}
\authorrunning{P. Lorenc et al.}
\institute{Faculty of Electrical Engineering, Czech Technical University in Prague, Czechia \\
\email{\{lorenpe2,gargitom,pichljan,konrajak,marekp17,kobzaond\}@fel.cvut.cz}
\and
Czech Institute of Informatics, Robotics and Cybernetics, Czech Technical University in Prague, Czechia \\
\email{jan.sedivy@cvut.cz}}
\maketitle              
\begin{abstract}
Conversational agents are usually designed for closed-world environments. Unfortunately, users can behave unexpectedly. Based on the open-world environment, we often encounter the situation that the training and test data are sampled from different distributions. Then, data from different distributions are called out-of-domain (OOD). A robust conversational agent needs to react to these OOD utterances adequately. Thus, the importance of robust OOD detection is emphasized. Unfortunately, collecting OOD data is a challenging task. We have designed an OOD detection algorithm independent of OOD data that outperforms a wide range of current state-of-the-art algorithms on publicly available datasets. Our algorithm is based on a simple but efficient approach of combining metric learning with adaptive decision boundary. Furthermore, compared to other algorithms, we have found that our proposed algorithm has significantly improved OOD performance in a scenario with a lower number of classes while preserving the accuracy for in-domain (IND) classes.

\end{abstract}
\section{Introduction}


Conversational interfaces built on top of Alexa or Siri mainly depend on dialogue management, which is responsible for coherent reacting to user utterances and sustaining the conversational flow \cite{conversation_agent_intro}. The dialogue management is typically based on Natural Language Understanding (NLU) classifications \cite{pichl2020alquist}. Nevertheless, based on the open-world assumption \cite{Keet2013_open_world_assumption}, the system cannot be prepared for all possible utterances. The utterances not taken from the train distribution are called out-of-domain (OOD). An example of a conversation with the critical necessity for OOD detection is shown in Table \ref{fig:example_of_ood_conversation2}. Therefore, we focus on an algorithm for OOD detection in conversational domain. Our algorithm is based on a simple but efficient approach of combining metric learning with adaptive decision boundary. To the best of our knowledge, we have not seen the proposed combination previously. Beside that, our algorithm also preserves performance for In-Domain (IND) classification as the two are usually performed together \cite{konrad2021alquist4}. Additionally, our algorithm does not require collecting OOD data (Out-of-domain Data Independent) and outperforms a wide range of current state-of-the-art algorithms for OOD detection on publicly available datasets. 






\begin{table}[t]\centering
\caption{Example of the difference between in-domain (IND) and out-of-domain (OOD) utterances for a geographical conversational agent}
\label{fig:example_of_ood_conversation2}
\begin{tabular}{rl}
User:  \,\,\,\, &  What is the population of Italy?    (IND)            \\
Agent: \,\,\,\, &  About 60 million. \\
User:  \,\,\,\, &  Great. Is there any news about the Italian prime minister? (OOD) \\
Agent: \,\,\,\, &  I am afraid that I cannot answer it. My knowledge is in geography!
\end{tabular}
\end{table}


\section{Related Work}

\subsubsection{Out-of-domain Data Dependent}

If we get access to OOD data specific to our IND classes, we can find a threshold that optimally separates IND and OOD examples as shown in \cite{larson2019evaluation_clinc_dataset}. The threshold is a trade-off between IND accuracy and OOD performance. The same paper shows that we can set OOD examples as $n+1$ class and train the classification model with other IND classes. The aforementioned approaches can be used on artificially created OOD instances from IND training examples \cite{odist} or enlarge known OOD training data \cite{chen2021gold} with the help of a pretrained language model. 

\subsubsection{Out-of-domain Data Independent}

The collection of OOD data is a resource-intensive process. Therefore, recent research also focuses on detecting OOD without the need to specify OOD data. 

An example of metric learning for OOD detection is in \cite{lin2019margin_loss}. They learn deep discriminative features by forcing the network to maximize interclass variance and minimize intraclass variance. Contrary to our approach, they learn features with a recurrent neural network and focus solely on OOD detection without focusing on IND performance.

The decision boundary for OOD was introduced in \cite{shu2017doc}. Their algorithm uses a post-processing step to find proper decision boundaries around IND classes. Contrary to our approach, they select the threshold for each IND class-based statistical distribution of the model's confidences. Another example of decision boundary is used in \cite{zhang2021adaptive_decision_boundary}. They proposed that the bounded spherical area greatly reduces the risk of treating OOD as IND in high-dimensional vector spaces. However, their method depends on fine-tuning BERT \cite{devlin2019bert}, which is computationally demanding \cite{reimers2019sentencebert}.

\section{Proposed Algorithm}

Let $D_{ID} = \{(x_1, y_1), ...,(x_n,y_n) \}$ be a dataset , where $x_i$ is vector representation of input utterance and $y_i \in T$ is its class. Then $T = \{C_1, ...C_i\}$ is a set of seen classes. Furthermore, $n_i$ is the number of examples for $i$-th class. 


The first step includes learning the transformation function $T(x)$, which maximizes interclass variance and minimizes intraclass variance. After application of $T(x)$ on every $x$ vector it increases the point density of each class $C_i$ around its centroid $c_i$:

\begin{equation} \label{eq:centroid}
c_i=\frac{1}{n_i} \sum_{x_i \in C_i} T(x_i)
\end{equation} 

where $n_i$ is the number of examples for $i$-th class $C_i$. 

The following step searches for decision boundary $r_i$ specific to each $i$-th class. To select best boundary $r_i$, we mark $i$-th class $C_i$ as IND class $C_{IND}$ and all other $C_s$, where $s \neq i$, as OOD class $C_{OOD}$. Then, for chosen $i$-th class $C_i$, we obtain the best threshold value $r_i$ balancing the $d(x_{IND}, c_i)$ and $d(x_{OOD}, c_i)$, where $d(x,y)$ is the normalized euclidean distance between vector $x$ and $y$, $x_{IND} \in C_{IND}$ and $x_{OOD} \in C_{OOD}$. Altogether, we define the stopping criterion $F(C_{IND}, C_{OOD}, r_i)$ as:
\begin{equation}
\begin{adjustbox}{width=0.9\columnwidth}
    $F(C_{IND}, C_{OOD}, r_i) =  \frac{\sum_{x\in C_{OOD}} (d(x, c_i) - r_i)}{\sum_{\forall s, s \neq i} n_s}  +  \frac{\sum_{x\in C_{IND}} (d(x, c_i) - r_i)}{n_i} * \beta_i$
\end{adjustbox}
\end{equation}
where $\beta_i$ is a hyper-parameter to normalize the importance of OOD performance. We have empirically observed that lower values of $\beta$ are better for lower numbers of IND classes and we suggest to use:

\begin{equation}
\beta_i ={\frac{\sum_{\forall s, s \neq i} n_s}{n_i}}
\end{equation}

for $i$-th IND class $C_{IND}$ to counter the imbalance between the number of OOD and IND examples.


Then, we minimize $max(F(C_{IND}, C_{OOD}, r_i), 0)$ by iteratively increasing the threshold value $r_i$ and evaluating the stopping criterion. We stop searching when we reach the minimum or when the number of steps exceeds the maximum iteration limit.




\section{Experiments}

This section introduces datasets, experimental setting and results. We also discuss ablation experiments.

\subsection{Datasets}
Following \cite{zhang2021adaptive_decision_boundary}, we used two publicly available datasets -- BANKING77 \cite{Casanueva2020_banking77} and CLINC150 \cite{larson2019evaluation_clinc_dataset}. The BANKING77 dataset contains 77 classes and 13,083 customer service queries.


The CLINC150 dataset contains 150 classes, 22,500 in-domain queries and 1,200 out-of-domain queries. 
An example of such queries is shown in Table \ref{tab:clinc150_example}.

\begin{table}[h]\centering
\caption{Example of customer service queries in CLINC150 \cite{larson2019evaluation_clinc_dataset}}
\label{tab:clinc150_example}
\begin{tabular}{l|l}
\textbf{Intent name} \,\,\,\, & \,\,\,\, \textbf{Utterance} \\ \hline
change\_speed \,\,\,\, & \,\,\,\,   will you please slow down your voice                \\
shopping\_list \,\,\,\, & \,\,\,\, show everything on my to buy list \\
OOD \,\,\,\, & \,\,\,\, what is the name of the 13th president       
\end{tabular}
\end{table}

\subsection{Experimental Setting}

Following \cite{zhang2021adaptive_decision_boundary}, we randomly select set of IND classes from the train set and integrate them into the test set as OOD. It results in various proportions between IND and OOD utterances -- 1:4 (25\% of IND and 75\% of OOD), 1:2 (50\% of IND and 50\% of OOD) and 3:4 (75\% of IND and 25\% of OOD). The accuracy and macro F1-score were computed as the average over 10 runs. 
The following pretrained sentence embeddings were used -- \textbf{Universal Sentence Encoder} (USE)\footnote{Deep Average Network (\textbf{DAN}) and with Transformer encoder (\textbf{TRAN})} by \cite{cer2018universal_sentence_encoder} and \textbf{Sentence-BERT} (SBERT) by \cite{reimers2019sentencebert}. 
As learning objectives for transformation function $T(x)$ we choose --- Triplet Loss \cite{hoffer2018deep_triplet_loss} and Large Margin Cosine Loss (LMCL) \cite{wang2018cosface}. Both learning objectives attempt to maximize interclass variance and minimize intraclass variance. All hyper-parameters of LMCL were set to values suggested by \cite{wang2018cosface}, and the hyper-parameters of Triplet Loss were set to the default value of its Tensorflow implementation. According to \cite{zhang2021adaptive_decision_boundary}, we compare our approach to several models: Maximum softmax probability \cite{baseline_msp} (MSP), OpenMax \cite{openMax}, Deep Open Classification \cite{shu2017doc} (DOC), Adaptive Decision Boundary \cite{zhang2021adaptive_decision_boundary} (ADB), and ODIST \cite{odist}. MSP calculates the softmax probability of known samples and rejects the samples with low confidence determined by threshold. OpenMax fits Weibull distribution to the outputs of the penultimate layer, but still needs negative samples for selecting the best hyper-parameters. DOC uses the sigmoid function and calculates the confidence threshold based on Gaussian statistics. ADB learns the adaptive spherical decision boundary for each known class with the aid of well-trained features. In addition, ODIST can create out-of-domain instances from the in-domain training examples with the help of a pre-trained language model. All computations were run on a virtual instance\footnote{AWS ml.m5.4xlarge}.





\subsection{Results}

Our measurement revealed a significant difference between different types of embeddings. USE-TRAN shows outstanding performance in all measurements. The combination of USE-TRAN with LMCL outperforms the current state-of-the-art approach in the majority of evaluations. We can also observe how different ratios of examples influence the performance of the models and conclude that our algorithm is more superior in scenarios with a lower number of IND classes. This can be beneficial for conversational agents like \cite{finch2020emora} or \cite{pichl2020alquist}.

\begin{table}[t]
\caption{Results on CLINC150 dataset. (1) - Results taken from \cite{zhang2021adaptive_decision_boundary} (2) - Results taken from \cite{odist}. (3) - Mean of own measurements based on USE-TRAN where $\pm$ is standard deviation}
\label{tab:clinc_result}
\centering
\begin{adjustbox}{width=1\textwidth}
\begin{tabular}{l|cc|cc|cc|c}
                            & \multicolumn{2}{c|}{25\% known ratio}                                                                                          & \multicolumn{2}{c|}{50\% known ratio}                                                                                          & \multicolumn{2}{c|}{75\% known ratio}                                                                                          &      \\ \hline
\textbf{Method}             & \multicolumn{1}{c}{\textbf{Accuracy}} & \multicolumn{1}{c|}{\textbf{F1}} & \multicolumn{1}{c}{\textbf{Accuracy}} & \multicolumn{1}{c|}{\textbf{F1}} & \multicolumn{1}{c}{\textbf{Accuracy}} & \multicolumn{1}{c|}{\textbf{F1}} & Note \\ \hline
\textbf{MSP}                & 47.02                                 & 47.62                                                                                  & 62.96                                 & 70.41                                                                                  & 74.07                                 & 82.38                                                                                  & (1)  \\
\textbf{DOC}                & 74.97                                 & 66.37                                                                                  & 77.16                                 & 78.26                                                                                  & 78.73                                 & 83.59                                                                                  & (1)  \\
\textbf{OpenMax}            & 68.50                                 & 61.99                                                                                  & 80.11                                 & 80.56                                                                                  & 76.80                                 & 73.16                                                                                  & (1)  \\
\textbf{DeepUnk}            & 81.43                                 & 71.16                                                                                  & 83.35                                 & 82.16                                                                                  & 83.71                                 & 86.23                                                                                  & (1)  \\
\textbf{ADB}                & 87.59                                 & 77.19                                                                                  & 86.54                                 & 85.05                                                                                  & 86.32                                 & 88.53                                                                                  & (1)  \\
\textbf{ODIST}              & 89.79                                 & UNK                                                                                    & 88.61                                 & UNK                                                                                    & 87.70                                 & UNK                                                                                    & (2)  \\ \hline
$\mathbf{Our}_\mathbf{LMCL}$    & \,\,  \textbf{91.81} $\pm$  0.11   \,\, &\,\,  \textbf{85.90} $\pm$  0.08  \,\,    & \,\, 88.81 $\pm$  0.15 \,\,  &\,\,  89.19 $\pm$  0.09  \,\,  & \,\, \textbf{88.54} $\pm$  0.05  \,\,  & \,\,  \textbf{92.21} $\pm$  0.10  \,\,  & (3)  \\
$\mathbf{Our}_\mathbf{Triplet}$ & \,\,  90.28 $\pm$  0.07  \,\,   & \,\, 84.82 $\pm$  0.14 \,\,    &\,\, \textbf{88.89} $\pm$  0.03 \,\,     & \,\, \textbf{89.44} $\pm$  0.04  \,\,   &\,\,  87.81 $\pm$  0.11   \,\,  &\,\,  91.72 $\pm$  0.17   \,\,  & (3) 
\end{tabular}
\end{adjustbox}
\end{table}

\begin{table}[t]
\caption{Results on BANKING77 dataset. (1) - Results taken from \cite{zhang2021adaptive_decision_boundary} (2) - Results taken from \cite{odist}. (3) - Mean of own measurements based on USE-TRAN where $\pm$ is standard deviation}
\label{tab:banking_result}
\centering
\begin{adjustbox}{width=1\textwidth}
\begin{tabular}{l|cc|cc|cc|c}
                            & \multicolumn{2}{c|}{25\% known ratio}       & \multicolumn{2}{c|}{50\% known ratio}       & \multicolumn{2}{c|}{75\% known ratio}       &      \\ \hline
\textbf{Method}             & \textbf{Accuracy}    & \textbf{F1}          & \textbf{Accuracy}    & \textbf{F1}          & \textbf{Accuracy}    & \textbf{F1}          & Note \\ \hline
\textbf{MSP}                & 43.67                & 50.09                & 59.73                & 71.18                & 75.89                & 83.60                & (1)  \\
\textbf{DOC}                & 56.99                & 58.03                & 64.81                & 73.12                & 76.77                & 83.34                & (1)  \\
\textbf{OpenMax}            & 49.94                & 54.14                & 65.31                & 74.24                & 77.45                & 84.07                & (1)  \\
\textbf{DeepUnk}            & 64.21                & 61.36                & 72.73                & 77.53                & 78.52                & 84.31                & (1)  \\
\textbf{ADB}                & 78.85                & 71.62                & 78.86                & 80.90                & 81.08                & 85.96                & (1)  \\
\textbf{ODIST}              & 81.69                &      UNK                 & 80.90                &              UNK        & 82.79                &                 UNK     & (2)  \\ \hline
$\mathbf{Our}_\mathbf{LMCL}$    & \,\,  \textbf{85.71} $\pm$ 0.13 \,\, &\,\,   \textbf{78.86} $\pm$ 0.10 \,\,  & \,\,  \textbf{83.78} $\pm$  0.14 \,\, & \,\,  \textbf{84.93} $\pm$  0.08 \,\, & \,\,  \textbf{84.40} $\pm$  0.21 \,\, & \,\,    \textbf{88.39} $\pm$  0.11 \,\, & (3)  \\
$\mathbf{Our}_\mathbf{Triplet}$ & \,\,  82.71 $\pm$  0.34 \,\, & \,\, 70.02  $\pm$  0.18 \,\, & \,\,  81.83 $\pm$  0.15 \,\,  & \,\,    83.07 $\pm$  0.15 \,\, & \,\,   81.82 $\pm$  0.08 \,\,  & \,\,   86.94 $\pm$  0.09 \,\, & (3) 
\end{tabular}
\end{adjustbox}
\end{table}

\begin{table}[b]
\caption{Results on CLINC150 dataset. (1) - Results taken from \cite{zhang2021adaptive_decision_boundary} (2) - Results taken from \cite{odist}. (3) - Own measurement with different embeddings (USE-DAN / USE-TRAN / SBERT).}
\label{tab:clinc_result_add}
\centering
\begin{adjustbox}{width=1\textwidth}
\begin{tabular}{l|cc|cc|cc|c}
                            & \multicolumn{2}{c|}{25\% known ratio}       & \multicolumn{2}{c|}{50\% known ratio}       & \multicolumn{2}{c|}{75\% known ratio}       &      \\ \hline
\textbf{Method}             & \textbf{\begin{tabular}[c]{@{}c@{}}F1\\  (OOD)\end{tabular}} & \textbf{\begin{tabular}[c]{@{}c@{}}F1\\  (IND)\end{tabular}} & \textbf{\begin{tabular}[c]{@{}c@{}}F1\\  (OOD)\end{tabular}} & \textbf{\begin{tabular}[c]{@{}c@{}}F1\\  (IND)\end{tabular}} & \textbf{\begin{tabular}[c]{@{}c@{}}F1\\  (OOD)\end{tabular}} & \textbf{\begin{tabular}[c]{@{}c@{}}F1\\  (IND)\end{tabular}}    & Note \\ \hline
\textbf{MSP}                & 50.88                & 47.53                & 57.62                & 70.58                & 59.08                & 82.59                & (1)  \\
\textbf{DOC}                & 81.98                & 65.96                & 79.00                & 78.25                & 72.87                & 83.69                & (1)  \\
\textbf{OpenMax}            & 75.76                & 61.62                & 81.89                & 80.54                & 76.35                & 73.13                & (1)  \\
\textbf{DeepUnk}            & 87.33                & 70.73                & 85.85                & 82.11                & 81.15                & 86.27                & (1)  \\
\textbf{ADB}                & 91.84                & 76.80                & 88.65                & 85.00                & 83.92                & 88.58                & (1)  \\
\textbf{ODIST}              & 93.42                & 79.69                & \textbf{90.62}       & 86.52                & \textbf{85.86}       & 89.33                & (2)  \\ \hline
$\mathbf{Our}_\mathbf{LMCL}$    & \,\,93.2 /  \textbf{94.5} /  92.7 \,\,&\,\, 83.0 /  \textbf{85.6} /  81.0 \,\, &\,\, 86.5 /  88.9 /  85.2 \,\,&\,\, 86.2 /  89.2 /  83.6 \,\,& \,\,74.7 /  78.4 /  71.4 \,\,&\,\, 90.0 /  \textbf{92.3} /  87.7 \,\,& (3)  \\
$\mathbf{Our}_\mathbf{Triplet}$ & \,\,91.6 /  93.3 /  91.1 \,\,&\,\, 80.7 /  84.6 /  78.8 \,\,&\,\, 84.6 /  89.0 /  84.2 \,\,& \,\, 85.3 /  \textbf{89.4} /  84.1 \,\, & \,\, 70.2 /  76.6 /  69.2 \,\, & \,\, 89.0 /  91.8 /  87.2 \,\, & (3) 
\end{tabular}
\end{adjustbox}

\end{table}

\begin{table}[t]
\caption{Results on BANKING77 dataset. (1) - Results taken from \cite{zhang2021adaptive_decision_boundary} (2) - Results taken from \cite{odist}. (3) - Own measurement with different embeddings (USE-DAN / USE-TRAN / SBERT).}
\label{tab:banking_result_add}
\centering
\begin{adjustbox}{width=1\textwidth}
\begin{tabular}{l|cc|cc|cc|c}
                            & \multicolumn{2}{c|}{25\% known ratio}       & \multicolumn{2}{c|}{50\% known ratio}       & \multicolumn{2}{c|}{75\% known ratio}       &      \\ \hline
\textbf{Method}             & \textbf{\begin{tabular}[c]{@{}c@{}}F1\\  (OOD)\end{tabular}} & \textbf{\begin{tabular}[c]{@{}c@{}}F1\\  (IND)\end{tabular}} & \textbf{\begin{tabular}[c]{@{}c@{}}F1\\  (OOD)\end{tabular}} & \textbf{\begin{tabular}[c]{@{}c@{}}F1\\  (IND)\end{tabular}} & \textbf{\begin{tabular}[c]{@{}c@{}}F1\\  (OOD)\end{tabular}} & \textbf{\begin{tabular}[c]{@{}c@{}}F1\\  (IND)\end{tabular}}    & Note \\ \hline
\textbf{MSP}                & 41.43                & 50.55                & 41.19                & 71.97                & 39.23                & 84.36                & (1)  \\
\textbf{DOC}                & 61.42                & 57.85                & 55.14                & 73.59                & 50.60                & 83.91                & (1)  \\
\textbf{OpenMax}            & 51.32                & 54.28                & 54.33                & 74.76                & 50.85                & 84.64                & (1)  \\
\textbf{DeepUnk}            & 70.44                & 60.88                & 69.53                & 77.74                & 58.54                & 84.75                & (1)  \\
\textbf{ADB}                & 84.56                & 70.94                & 78.44                & 80.96                & 66.47                & 86.29                & (1)  \\
\textbf{ODIST}              & 87.11                & 72.72                & 81.32       & 81.79                & 71.95       & 87.20                & (2)  \\ \hline
$\mathbf{Our}_\mathbf{LMCL}$    & \,\, 87.5 /  \textbf{89.9} /  89.7 \,\, & \,\, 74.6 /  \textbf{78.4} /  76.7 \,\, & \,\, 82.4 /  \textbf{83.9} /  82.9 \,\, & \,\, 82.9 /  \textbf{84.9} /  82.6 \,\, & \,\, 67.0 /  \textbf{73.1} /  69.3 \,\, & \,\, 86.9 /  \textbf{88.7} /  87.0 \,\, & (3)  \\
$\mathbf{Our}_\mathbf{Triplet}$ & \,\, 87.5 /  88.0 /  87.3 \,\, & \,\, 66.8 /  69.1 /  72.9 \,\, & \,\, 77.5 /  81.9 /  81.2 \,\, & \,\, 64.6 /  83.0 /  82.1 \,\, & \,\, 64.5 /  66.8 /  65.9 \,\, & \,\, 84.7 /  87.2 /  85.4 \,\, & (3) 
\end{tabular}
\end{adjustbox}

\end{table}

\subsection{Ablation study}

\begin{figure}[]
\centering
\includegraphics[width=.5\textwidth]{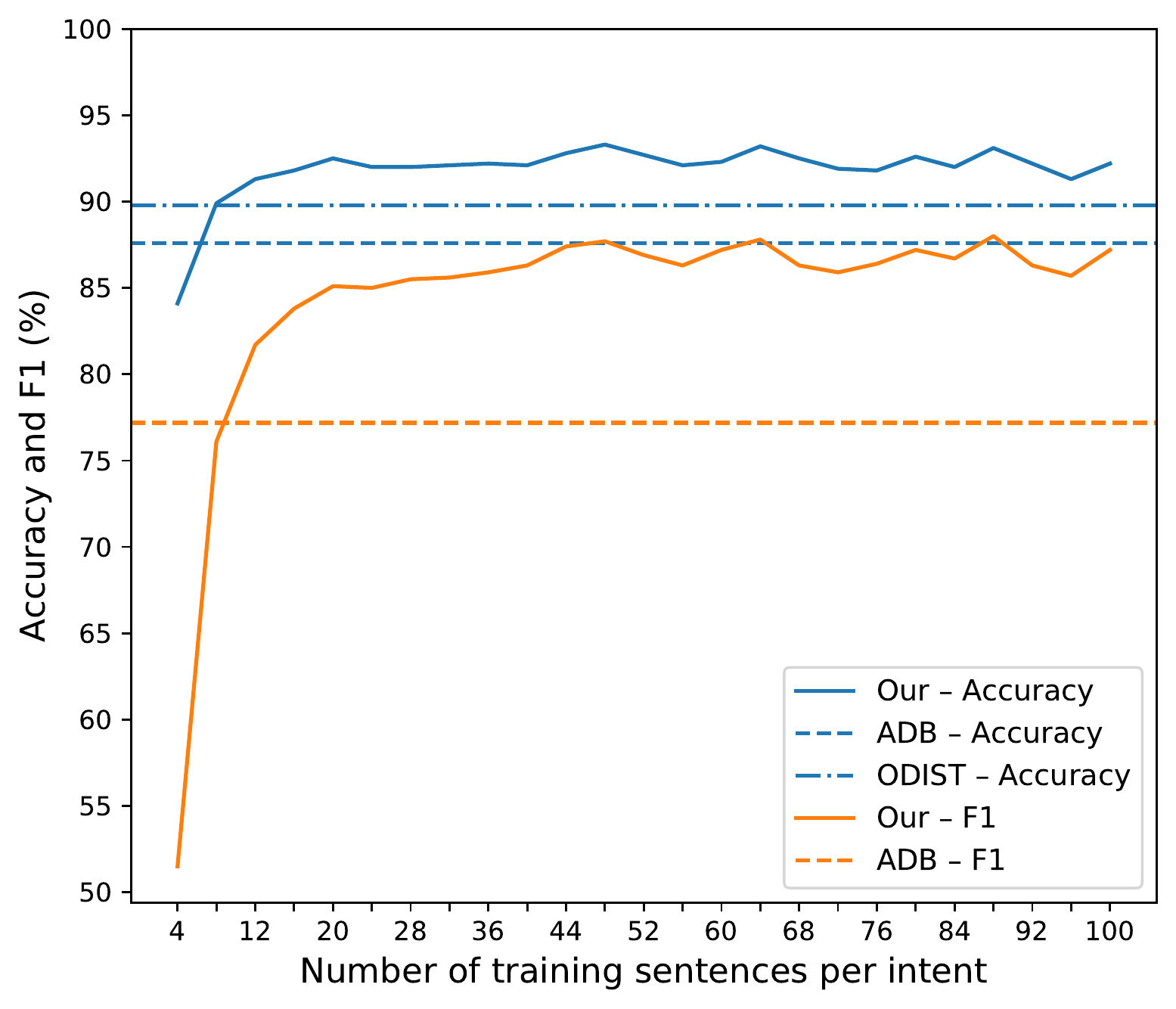}
\caption{Influence of the number of training sentences on the accuracy of the CLINC150 dataset. Only 25\% of classes are taken as IND}
\label{fig:num_train_examples}

\end{figure}

\textbf{How does the number of examples influence the performance?} -- Since the collection of IND training data can be an expensive process, we evaluate the performance of our proposed algorithm under a scenario with a limited number of training examples. Our evaluation focused on accuracy and macro F1-score over all classes. We compare our results with the best performing results of ADB and ODIST. The results indicate that our method is very efficient even with a small fraction of train data. The results shown in Figure \ref{fig:num_train_examples} were performed with a random limited selection of train sentences in the CLINC150 dataset. Shown performance is the average from 5 runs.

\textbf{What is the difference in performance between IND and OOD?} Similar to \cite{zhang2021adaptive_decision_boundary}, we also evaluate F1 over OOD and over IND, respectively. The results shown in Table \ref{tab:clinc_result_add} and Table \ref{tab:banking_result_add} have demonstrated the superior performance of our algorithm in a scenario where we classify into a smaller number of classes. This scenario is typical for conversational agents \cite{konrad2021alquist4}.




\section{Conclusion}
In summary, we proposed a novel algorithm for OOD detection with a combination of metric learning and adaptive boundary. The present findings might help solve the problem of OOD detection without the need for OOD training data. The theoretical part is focused on metric learning and the creation of adaptive boundaries, the crucial parts of our two-step algorithm. We describe our improvements over existing approaches and introduce our novel stopping criterion. The strengths of our work are then verified in a controlled experiment. The comparison on different datasets shows that the algorithm achieves superior performance over other state-of-the-art approaches. We showed that our algorithm finds the best decision boundary concerning IND accuracy and OOD performance, together with feasible computational requirements. We release all source code to reproduce our results \footnote{https://github.com/tgargiani/Adaptive-Boundary}.

\subsection{Future work and usage}

There is a rising group of other possibilities for loss functions used in the metric learning step. The Circle loss \cite{sun2020circleloss} or Quadruplet Loss \cite{chen2017quadruplettriplet} represent the next possible improvement of our algorithm. They show superior performance over LMCL. We want to investigate if improved metric learning will also increase the performance of our algorithm.  
With that in mind, we propose that our suggestion for a two-step algorithm is ready for usage in many scenarios concerning user input, such as a conversational agent or searching for information over a finite set of documents. In all these situations, there is a possibility that the user query is impossible to answer, leading to the fallback scenario.

\subsubsection{Acknowledgments.}

This research was partially supported by the Grant Agency of the Czech Technical University in Prague, grant (SGS22/082/OHK3/1T/37).

%
%
%
%
\bibliographystyle{splncs04}
\bibliography{custom}
\end{document}